\documentclass[10pt,a4paper,twocolumn]{article}
\usepackage[a4paper,left=1.45cm,right=1.45cm,top=1.6cm,bottom=1.6cm,columnsep=0.36cm]{geometry}
\usepackage{times}
\usepackage{graphicx}
\usepackage{amsmath,amssymb}
\usepackage{booktabs}
\usepackage{caption}
\usepackage{titlesec}
\usepackage{fancyhdr}
\usepackage{enumitem}
\usepackage{xcolor}
\usepackage{tikz}
\usetikzlibrary{arrows.meta,positioning}
\titleformat{\section}{\bfseries\fontsize{10.8}{11.8}\selectfont}{\thesection}{0.45em}{}
\titleformat{\subsection}{\itshape\fontsize{9.6}{10.5}\selectfont}{\thesubsection}{0.45em}{}
\titlespacing*{\section}{0pt}{0.42em}{0.18em}
\titlespacing*{\subsection}{0pt}{0.32em}{0.10em}
\setlist[itemize]{leftmargin=1.0em,itemsep=0pt,topsep=1pt,parsep=0pt}

\begin{document}
\twocolumn[
\begin{center}
{\fontsize{15.3}{17}\selectfont\bfseries
MambaPSA: A Mamba-based Replacement for C2PSA in YOLO26
\par}
\vspace{0.30em}
Sheng-Wei Chan$^{1}$, Chia-Min Lin$^{1}$, Hsin-Jui Pan$^{1}$, Ching-Yu Tsai$^{1}$, Chih-Hsiang Yang$^{1}$, Yung-Che Wang$^{1}$, Jen-Shiun Chiang$^{1*}$\\
{\small $^{1}$Department of Electrical and Computer Engineering, Tamkang University, New Taipei City, Taiwan\\
$^{*}$chiang@mail.tku.edu.tw}
\vspace{0.50em}
\end{center}
\vspace{0.45em}

\noindent\textbf{Abstract}--- State space models (SSMs), notably Mamba, have recently emerged as efficient alternatives to self-attention with linear computational complexity. We investigate the integration of Mamba into YOLO26, the latest non-maximum suppression (NMS)-free object detection framework, by proposing \textit{MambaPSA}---a lightweight Mamba-based replacement for the C2PSA block at the end of the backbone. To complement this study, we additionally insert a bidirectional Vision Mamba (BiViM) module at the P3, P4, and P5 levels of the neck. Experiments on PASCAL VOC 2007+2012 show that MambaPSA reduces parameters by 2.9\%, FLOPs by 12.1\%, and improves CPU inference throughput by 17.6\% (from 17 to 20 FPS) with negligible accuracy change ($-0.1$ mAP$_{50:95}$), while the P4 BiViM placement yields the best accuracy gain ($+0.9$ mAP$_{50:95}$). These results suggest that SSMs offer a favorable efficiency--accuracy trade-off when replacing attention-based blocks in NMS-free lightweight detectors.
\vspace{0.6em}
]

\begin{table*}[!t]
\centering
\caption{Per-class mAP$_{50:95}$ (\%) on PASCAL VOC 2007 test. Best per class is in bold. Methods: Base = YOLO26-n baseline; +P$k$ BiViM = BiVisionMamba inserted at the P$k$ level of the neck; MPSA = MambaPSA (ours). Class abbreviations: aero=aeroplane, bike=bicycle, bot=bottle, cha=chair, tab=diningtable, hor=horse, mbk=motorbike, per=person, pot=pottedplant, she=sheep, sof=sofa, tra=train, tv=tvmonitor.}
\label{tab:perclass}
\footnotesize
\setlength{\tabcolsep}{2.5pt}
\begin{tabular}{@{}lccccccccccccccccccccc@{}}
\toprule
Method & aero & bike & bird & boat & bot & bus & car & cat & cha & cow & tab & dog & hor & mbk & per & pot & she & sof & tra & tv & all \\
\midrule
Base & 55.6 & 59.3 & 40.0 & \textbf{39.1} & 30.5 & 68.6 & 64.9 & 60.9 & 30.8 & 47.1 & 48.8 & 54.8 & 59.0 & 55.9 & \textbf{53.5} & 19.3 & 48.3 & 51.4 & 60.4 & \textbf{50.6} & 49.9 \\
+ P3 BiViM & 54.1 & 57.3 & 37.2 & 36.3 & 29.5 & 67.8 & 64.4 & 58.9 & 28.5 & 48.6 & 47.3 & 52.8 & 56.9 & 55.0 & 52.6 & 17.9 & 46.9 & 48.1 & 59.8 & 47.9 & 48.4 \\
+ P4 BiViM & \textbf{58.1} & 60.0 & 41.0 & 37.2 & \textbf{31.2} & \textbf{70.7} & 65.4 & \textbf{62.8} & \textbf{31.5} & \textbf{48.6} & 49.8 & 54.8 & 59.0 & \textbf{57.1} & 52.9 & \textbf{21.5} & 48.5 & 52.5 & \textbf{62.8} & 49.9 & \textbf{50.8} \\
+ P5 BiViM & 56.0 & \textbf{61.4} & \textbf{42.9} & 37.3 & 30.0 & 70.5 & \textbf{65.5} & 60.3 & 31.4 & 46.1 & \textbf{51.6} & \textbf{55.8} & \textbf{60.8} & 56.5 & 53.2 & 17.9 & 49.1 & \textbf{53.7} & 61.1 & 49.9 & 50.6 \\
\textbf{MPSA (ours)} & 56.6 & 58.9 & 42.5 & 37.0 & 29.7 & 70.4 & 65.3 & 60.7 & 29.7 & 47.1 & 45.6 & 54.8 & 59.1 & 55.1 & 52.8 & 18.9 & \textbf{50.1} & 51.6 & 58.5 & \textbf{50.6} & 49.8 \\
\bottomrule
\end{tabular}
\end{table*}

\section{Introduction}
The YOLO family of single-stage detectors~\cite{yolo,ultralytics} continuously evolves through architectural refinements. YOLO26, released in January 2026~\cite{ultralytics}, adopts a non-maximum suppression (NMS)-free dual-head training strategy and simplifies the regression head. At the end of its backbone, YOLO26 employs the C2PSA block, which performs global feature aggregation through a position-sensitive self-attention (PSA) mechanism built on the cross-stage partial (CSP) topology~\cite{cspnet}. While effective, self-attention~\cite{attention,vit} scales quadratically with token count, constraining its use in lightweight edge deployments.

State space models (SSMs), notably Mamba~\cite{mamba}, recently emerged as efficient sequence models with linear complexity through a selective scan mechanism, building on structured SSMs such as S4~\cite{s4}. Vision adaptations such as Vision Mamba~\cite{vim} and VMamba~\cite{vmamba} have shown that SSMs rival Transformer-based backbones on classification, and recent works integrate Mamba into the YOLO family~\cite{mambayolo}. To our knowledge, however, whether Mamba can directly replace the C2PSA block in YOLO26 has not been examined in the public literature; we therefore propose \textit{MambaPSA}, a Mamba-based replacement that preserves the CSP wrapper. We additionally insert a bidirectional Vision Mamba (BiViM) module at three neck levels (P3, P4, P5) to identify effective positions for SSM-based feature aggregation. Our contributions are:
\begin{itemize}
\item We propose MambaPSA, a lightweight Mamba-based replacement for the C2PSA block of YOLO26.
\item We conduct a placement study of SSM modules at P3, P4, and P5 of the neck, revealing a non-monotonic relationship between placement and accuracy.
\item Empirical evaluation on PASCAL VOC shows that MambaPSA reduces FLOPs by 12.1\% and improves CPU inference throughput by 17.6\% with negligible accuracy degradation, while P4 BiViM yields the best accuracy gain.
\end{itemize}

\section{Related Work}
Mamba~\cite{mamba} introduces selective scanning that enables SSMs to model long sequences in linear time, building on S4~\cite{s4}. Vision Mamba~\cite{vim} and VMamba~\cite{vmamba} extend SSMs to image data via bidirectional or four-directional scan strategies. Several recent works integrate SSMs into YOLO detectors: Mamba-YOLO~\cite{mambayolo} replaces the C2f block in the PAFPN neck of YOLOv8, and subsequent works target the C3k2 block in YOLO11. The integration of SSMs with the C2PSA block of YOLO26---an NMS-free framework---remains, to our knowledge, unexplored.

\section{Method}
\begin{figure}[!t]
\centering
\begin{tikzpicture}[
  every node/.style={font=\scriptsize, align=center, inner sep=1.6pt},
  blk/.style={draw=gray!60!black, rounded corners=2pt, minimum height=4.0mm, minimum width=1.05cm, line width=0.45pt, fill=white},
  blkhi/.style={draw=red!75!black, rounded corners=2pt, minimum height=4.0mm, minimum width=1.45cm, line width=0.8pt, fill=red!15, font=\scriptsize\bfseries},
  arr/.style={-{Stealth[length=1.4mm,width=1.2mm]}, line width=0.5pt, draw=gray!55!black},
  arrhi/.style={-{Stealth[length=1.4mm,width=1.2mm]}, line width=0.65pt, draw=red!75!black, dashed},
]
% --- Section background containers ---
\fill[blue!7, rounded corners=4pt] (-0.80,-4.05) rectangle (0.80,0.50);
\draw[blue!40!black, rounded corners=4pt, line width=0.3pt, dashed] (-0.80,-4.05) rectangle (0.80,0.50);
\fill[yellow!15, rounded corners=4pt] (1.50,-4.05) rectangle (4.75,0.50);
\draw[orange!50!black, rounded corners=4pt, line width=0.3pt, dashed] (1.50,-4.05) rectangle (4.75,0.50);
\fill[green!8, rounded corners=4pt] (5.00,-4.05) rectangle (6.30,0.50);
\draw[green!40!black, rounded corners=4pt, line width=0.3pt, dashed] (5.00,-4.05) rectangle (6.30,0.50);

% --- Section labels ---
\node[font=\bfseries\footnotesize, text=blue!60!black] at (0,0.28) {Backbone};
\node[font=\bfseries\footnotesize, text=orange!65!black] at (3.12,0.28) {Neck (PAN-FPN)};
\node[font=\bfseries\footnotesize, text=green!50!black] at (5.65,0.28) {Head};

% --- Backbone column (x=0) ---
\node[blk] (in)  at (0, 0.00) {Input};
\node[blk] (stm) at (0,-0.50) {Stem};
\node[blk] (b2)  at (0,-1.00) {C3k2 P2};
\node[blk] (b3)  at (0,-1.50) {C3k2 P3};
\node[blk] (b4)  at (0,-2.00) {C3k2 P4};
\node[blk] (b5)  at (0,-2.50) {C3k2 P5};
\node[blk] (sp)  at (0,-3.00) {SPPF};
\node[blkhi] (mp) at (0,-3.60) {MambaPSA};

% --- Neck N3/N4/N5 (x=2.25) ---
\node[blk] (n3) at (2.25,-1.50) {N3};
\node[blk] (n4) at (2.25,-2.00) {N4};
\node[blk] (n5) at (2.25,-3.60) {N5};

% --- BiViM (x=3.85) ---
\node[blkhi] (v3) at (3.85,-1.50) {BiViM P3};
\node[blkhi] (v4) at (3.85,-2.00) {BiViM P4};
\node[blkhi] (v5) at (3.85,-3.60) {BiViM P5};

% --- Detect (x=5.65) ---
\node[blk] (d3) at (5.65,-1.50) {Detect};
\node[blk] (d4) at (5.65,-2.00) {Detect};
\node[blk] (d5) at (5.65,-3.60) {Detect};

% Arrows backbone chain
\draw[arr] (in)  -- (stm);
\draw[arr] (stm) -- (b2);
\draw[arr] (b2)  -- (b3);
\draw[arr] (b3)  -- (b4);
\draw[arr] (b4)  -- (b5);
\draw[arr] (b5)  -- (sp);
\draw[arr] (sp)  -- (mp);

% Backbone -> Neck
\draw[arr] (b3.east) -- (n3.west);
\draw[arr] (b4.east) -- (n4.west);
\draw[arr] (mp.east) -- (n5.west);

% Neck -> BiViM (red dashed)
\draw[arrhi] (n3.east) -- (v3.west);
\draw[arrhi] (n4.east) -- (v4.west);
\draw[arrhi] (n5.east) -- (v5.west);

% BiViM -> Detect
\draw[arr] (v3.east) -- (d3.west);
\draw[arr] (v4.east) -- (d4.west);
\draw[arr] (v5.east) -- (d5.west);

\end{tikzpicture}
\caption{YOLO26 with two SSM integration points (red). \textbf{MambaPSA} replaces the original C2PSA block at the end of the backbone. A bidirectional Vision Mamba block (\textbf{BiViM}, dashed red) is optionally inserted at P3, P4, or P5 of the neck; only one is active per variant. N3/N4/N5 denote neck features at the corresponding resolutions.}
\label{fig:arch}
\end{figure}
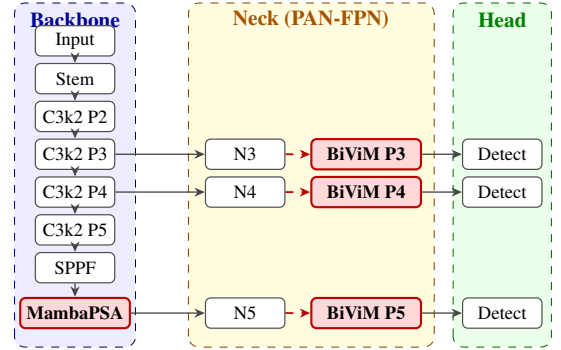

Fig.~\ref{fig:arch} illustrates the overall YOLO26 pipeline with the two SSM integration points, and Fig.~\ref{fig:blocks} details the internal structure of the two proposed blocks, MambaPSA and BiViM.

\subsection{MambaPSA}
MambaPSA preserves the CSP wrapper of C2PSA but replaces the internal self-attention branch with a Mamba core (Fig.~\ref{fig:blocks}, left). Given an input $\mathbf{x} \in \mathbb{R}^{C \times H \times W}$, a $1\times1$ convolution projects $\mathbf{x}$ to $2C'$ channels split into halves $\mathbf{a}$ and $\mathbf{b}$. The first half $\mathbf{a}$ is flattened into $HW$ tokens, processed by a single Mamba block~\cite{mamba}, and reshaped back; the second half $\mathbf{b}$ is preserved through identity. The two halves are concatenated and projected back to $C$ channels by a second $1\times1$ convolution. The Mamba core is configured as lightweight ($d_{\text{state}}{=}8$, $e{=}1$, mono-directional scan) so that the overall block is approximately parameter-neutral relative to C2PSA.

\subsection{Mamba in the Neck}
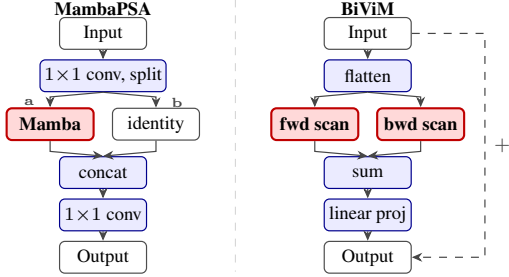
\begin{figure}[!t]
\centering
\begin{tikzpicture}[
  every node/.style={font=\scriptsize, align=center, inner sep=1.5pt},
  nd/.style={draw=gray!60!black, rounded corners=2pt, minimum height=4.2mm, minimum width=1.15cm, line width=0.45pt, fill=white},
  ndb/.style={draw=blue!55!black, rounded corners=2pt, minimum height=4.2mm, minimum width=1.15cm, line width=0.45pt, fill=blue!8},
  ndhi/.style={draw=red!75!black, rounded corners=2pt, minimum height=4.2mm, minimum width=1.15cm, line width=0.8pt, fill=red!15, font=\scriptsize\bfseries},
  a/.style={-{Stealth[length=1.4mm,width=1.2mm]}, line width=0.45pt, draw=gray!55!black},
  ares/.style={-{Stealth[length=1.4mm,width=1.2mm]}, line width=0.45pt, draw=gray!55!black, dashed},
]
% ===================== LEFT: MambaPSA =====================
\node[font=\bfseries\scriptsize] at (0,0.32) {MambaPSA};
\node[nd]   (li)  at (0, 0.00) {Input};
\node[ndb]  (lc1) at (0,-0.56) {$1{\times}1$ conv, split};
\node[ndhi] (la)  at (-0.7,-1.20) {Mamba};
\node[nd]   (lb)  at (0.7,-1.20) {identity};
\node[ndb]  (lcat)at (0,-1.84) {concat};
\node[ndb]  (lc2) at (0,-2.40) {$1{\times}1$ conv};
\node[nd]   (lo)  at (0,-2.96) {Output};

\draw[a] (li) -- (lc1);
\draw[a] (lc1.south) -- (-0.7,-0.84) -- (la.north);
\draw[a] (lc1.south) -- (0.7,-0.84) -- (lb.north);
\node[font=\tiny, text=gray!55!black] at (-0.98,-0.91) {$\mathbf{a}$};
\node[font=\tiny, text=gray!55!black] at (0.98,-0.91) {$\mathbf{b}$};
\draw[a] (la.south) -- (-0.7,-1.56) -- (lcat.north);
\draw[a] (lb.south) -- (0.7,-1.56) -- (lcat.north);
\draw[a] (lcat) -- (lc2);
\draw[a] (lc2) -- (lo);

% ===================== divider =====================
\draw[gray!35, line width=0.3pt, dashed] (1.75,0.45) -- (1.75,-3.15);

% ===================== RIGHT: BiViM =====================
\node[font=\bfseries\scriptsize] at (3.5,0.32) {BiViM};
\node[nd]   (ri)  at (3.5, 0.00) {Input};
\node[ndb]  (rf)  at (3.5,-0.56) {flatten};
\node[ndhi] (rfw) at (2.8,-1.20) {fwd scan};
\node[ndhi] (rbw) at (4.2,-1.20) {bwd scan};
\node[ndb]  (rsum)at (3.5,-1.84) {sum};
\node[ndb]  (rlin)at (3.5,-2.40) {linear proj};
\node[nd]   (ro)  at (3.5,-2.96) {Output};

\draw[a] (ri) -- (rf);
\draw[a] (rf.south) -- (2.8,-0.84) -- (rfw.north);
\draw[a] (rf.south) -- (4.2,-0.84) -- (rbw.north);
\draw[a] (rfw.south) -- (2.8,-1.56) -- (rsum.north);
\draw[a] (rbw.south) -- (4.2,-1.56) -- (rsum.north);
\draw[a] (rsum) -- (rlin);
\draw[a] (rlin) -- (ro);
% residual: input -> output (far right dashed)
\draw[ares] (ri.east) -- (5.05,0.0) -- (5.05,-2.96) -- (ro.east);
\node[font=\scriptsize, text=gray!55!black] at (5.28,-1.48) {$+$};

\end{tikzpicture}
\caption{Internal structure of the two proposed blocks. \textbf{Left:} MambaPSA splits the projected feature into two halves; half $\mathbf{a}$ passes through a single (mono-directional) Mamba core (red) while half $\mathbf{b}$ is kept as identity, before concatenation and a second $1{\times}1$ convolution. \textbf{Right:} BiViM scans the flattened tokens in forward and backward directions (red), sums the two outputs, applies a linear projection, and adds a residual connection.}
\label{fig:blocks}
\end{figure}

For the placement study, a bidirectional Vision Mamba (BiViM) block is applied immediately after the upsample-concat-C3k2 stage at the P3, P4, or P5 level (Fig.~\ref{fig:blocks}, right). The block scans the flattened token sequence in both directions, sums the two outputs, applies a linear projection, and adds a residual connection ($d_{\text{state}}{=}16$, $e{=}2$).

\section{Experiments}
\subsection{Setup}
We evaluate on PASCAL VOC. The training set combines VOC 2007 and 2012 trainval (16{,}551 images); evaluation is on VOC 2007 test (4{,}952 images, 20 classes)~\cite{voc}. All models are trained for 100 epochs using AdamW (lr $10^{-3}$, batch 32, image size $640{\times}640$) with a 5-epoch warmup, last-15-epoch mosaic close, and mixed-precision (fp16). Training is on a single NVIDIA RTX 4090; parameters are reported post-fusion. CPU throughput is measured on an Intel Core i7 workstation CPU (batch size 1, single run).

\subsection{Main Results}
Table~\ref{tab:main} compares the YOLO26-Nano baseline, three placement variants, and the proposed MambaPSA. MambaPSA reduces parameters by 2.9\% and FLOPs by 12.1\% relative to the baseline with only a 0.1 mAP$_{50:95}$ decrease. Among the placement variants, P4 yields the highest accuracy gain ($+0.9$), P3 degrades accuracy substantially ($-1.5$), and P5 provides moderate improvement ($+0.7$) at a parameter overhead of $+43.8\%$. In terms of parameter cost per $1\%$ \emph{relative} mAP$_{50:95}$ change (relative parameter change divided by relative mAP change), P4 is the most efficient ($+5.3\%$), P5 is over five times more expensive ($+31.3\%$), whereas MambaPSA reduces parameters rather than expending them ($-14.5\%$), placing it on the opposite side of the efficiency--accuracy trade-off.

On CPU inference, MambaPSA reaches 20 FPS versus the baseline's 17 FPS, a 17.6\% throughput improvement broadly consistent with the FLOP reduction. Precise mean $\pm$ std across multiple runs, alternative deployment runtimes (ONNX Runtime, OpenVINO), and edge-device profiling are deferred to future work.

\begin{table}[!t]
\centering
\caption{Comparison of MambaPSA and SSM placement variants against the YOLO26-Nano baseline on PASCAL VOC 2007 test. Values in parentheses are relative changes with respect to the baseline. CPU FPS is measured on an Intel Core i7 workstation CPU (batch size 1) from a single benchmarking run; ``---'' denotes not measured, as placement variants target the accuracy trade-off rather than efficiency.}
\label{tab:main}
\footnotesize
\setlength{\tabcolsep}{2pt}
\begin{tabular}{@{}lrrrr@{}}
\toprule
Method & Params & GFLOPs & mAP$_{50:95}$ & CPU FPS \\
\midrule
YOLO26-n & 2.4M & 5.8 & 49.9 & 17 \\
~+ P3 BiViM & 2.45M (+2.1\%) & 6.5 (+12.1\%) & 48.4 ($-$1.5) & --- \\
~+ P4 BiViM & 2.63M (+9.6\%) & 6.2 (+6.9\%) & \textbf{50.8} (+0.9) & --- \\
~+ P5 BiViM & 3.45M (+43.8\%) & 6.1 (+5.2\%) & 50.6 (+0.7) & --- \\
\textbf{MambaPSA} & \textbf{2.33M} ($-$2.9\%) & \textbf{5.1} ($-$12.1\%) & 49.8 ($-$0.1) & \textbf{20} (+17.6\%) \\
\bottomrule
\end{tabular}
\end{table}

\subsection{Per-class Analysis}
Table~\ref{tab:perclass} reports per-class mAP$_{50:95}$ across the 20 PASCAL VOC categories. The P4 variant attains the best accuracy on 9 categories, including large objects (\textit{bus} +2.1, \textit{train} +2.4, \textit{aeroplane} +2.5) and small objects (\textit{bottle} +0.7, \textit{pottedplant} +2.2), supporting its role as a balanced sweet spot. The P5 variant excels on categories with rich global context (\textit{bird} +2.9, \textit{horse} +1.8, \textit{sofa} +2.3) but degrades on small objects such as \textit{pottedplant} ($-$1.4). The P3 variant regresses on nearly all categories, suggesting that one-dimensional scanning over high-resolution feature maps may disrupt the local spatial structure required for fine-grained detection. MambaPSA exhibits a category-dependent trade-off with gains on animals and large vehicles (\textit{bird} +2.5, \textit{bus} +1.8, \textit{sheep} +1.8) and losses on indoor and elongated structures (\textit{diningtable} $-$3.2, \textit{boat} $-$2.1, \textit{train} $-$1.9).

\section{Discussion and Conclusion}
We propose MambaPSA, a lightweight Mamba-based replacement for the C2PSA block in YOLO26, achieving $\sim$12\% FLOP reduction and 17.6\% CPU throughput improvement with negligible accuracy impact on PASCAL VOC. The placement study identifies P4 as a favorable position for SSM-based aggregation. All results reflect a single training seed; inter-seed variance on VOC-nano scale can approach the magnitudes of the per-variant differences reported here, so the observed gaps should be interpreted as directional trends. Future work will address multi-seed stability, the combination of MambaPSA with the P4 BiViM placement, precise CPU/edge-device profiling across deployment runtimes, and MS COCO validation.

\section*{Acknowledgements}
This research work is partially supported by National Science and Technology Council, Taiwan, under grant number 114-2221-E-032-011.


\begin{thebibliography}{10}
\footnotesize
\setlength{\itemsep}{-1.5pt}
\setlength{\parsep}{0pt}
\bibitem{yolo}
J. Redmon, S. Divvala, R. Girshick, and A. Farhadi,
``You only look once: Unified, real-time object detection,''
in \textit{Proc. IEEE Conf. Comput. Vis. Pattern Recognit. (CVPR)}, 2016, pp. 779--788.

\bibitem{ultralytics}
G. Jocher, A. Chaurasia, and J. Qiu,
``Ultralytics YOLO,'' GitHub, \texttt{https://github.com/ultralytics/ultralytics}, 2026.

\bibitem{cspnet}
C.-Y. Wang \textit{et al.},
``CSPNet: A new backbone that can enhance learning capability of CNN,''
in \textit{CVPR Workshops}, 2020, pp. 390--391.

\bibitem{attention}
A. Vaswani \textit{et al.},
``Attention is all you need,''
in \textit{Proc. Adv. Neural Inf. Process. Syst. (NeurIPS)}, 2017, pp. 5998--6008.

\bibitem{vit}
A. Dosovitskiy \textit{et al.},
``An image is worth 16x16 words: Transformers for image recognition at scale,''
in \textit{Proc. Int. Conf. Learn. Representations (ICLR)}, 2021.

\bibitem{s4}
A. Gu, K. Goel, and C. R\'{e},
``Efficiently modeling long sequences with structured state spaces,''
in \textit{ICLR}, 2022.

\bibitem{mamba}
A. Gu and T. Dao,
``Mamba: Linear-time sequence modeling with selective state spaces,''
\textit{arXiv preprint arXiv:2312.00752}, 2023.

\bibitem{vim}
L. Zhu \textit{et al.},
``Vision Mamba: Efficient visual representation learning with bidirectional state space model,''
in \textit{Proc. Int. Conf. Mach. Learn. (ICML)}, 2024.

\bibitem{vmamba}
Y. Liu \textit{et al.},
``VMamba: Visual state space model,''
in \textit{NeurIPS}, 2024.

\bibitem{mambayolo}
Z. Wang \textit{et al.},
``Mamba YOLO: SSMs-based YOLO for object detection,''
in \textit{Proc. AAAI Conf. Artif. Intell.}, 2025.

\bibitem{voc}
M. Everingham \textit{et al.},
``The PASCAL Visual Object Classes (VOC) challenge,''
\textit{Int. J. Comput. Vis.}, vol. 88, no. 2, pp. 303--338, 2010.

\end{thebibliography}
\end{document}